\pdfoutput=1
\def\year{2019}\relax
\documentclass[letterpaper]{article} 
\usepackage{aaai19}  
\usepackage{times}  
\usepackage{helvet}  
\usepackage{courier}  
\usepackage{url}  
\usepackage{graphicx}  

\usepackage{url}
\usepackage{epsfig}
\usepackage{amsmath}
\usepackage{amsfonts,amssymb}
\usepackage{subcaption}
\usepackage{graphicx}
\usepackage{bm}
\usepackage{soul}
\usepackage{xcolor}
\usepackage[group-separator={,},group-minimum-digits={3}]{siunitx}
\usepackage{multicol}
\usepackage{multirow}
\usepackage{algorithm}
\usepackage{algorithmicx}
\usepackage{algpseudocode}

\newcommand{\toprule}{\noalign{\hrule height .08em}}
\newcommand{\bottomrule}{\noalign{\hrule height .08em}}

\newcommand{\newcite}[1]{\citeauthor{#1}~(\citeyear{#1})}

\newcommand*{\affaddr}[1]{#1} 
\newcommand*{\affmark}[1][*]{\textsuperscript{#1}}
\newcommand*{\email}[1]{\texttt{#1}}

\begin{document}
%
\title{A Deep Reinforced Sequence-to-Set Model for Multi-Label Text Classification}

\author{Pengcheng Yang\affmark[1,2], Shuming Ma\affmark[2], Yi Zhang\affmark[2], Junyang Lin\affmark[2,3], Qi Su\affmark[3], Xu Sun\affmark[1,2]\\
\affaddr{\affmark[1]Deep Learning Lab, Beijing Institute of Big Data Research, Peking University}\\
\affaddr{\affmark[2]MOE Key Lab of Computational Linguistics, School of EECS, Peking University}\\
\affaddr{\affmark[3]School of Foreign Languages, Peking University}\\
\email{\{yang\_pc, shumingma, zhangyi16, linjunyang, sukia, xusun\}@pku.edu.cn}\\
}

\maketitle

\begin{abstract}
Multi-label text classification (MLTC) aims to assign multiple labels to each sample in the dataset. The labels usually have internal correlations. However, traditional methods tend to ignore the correlations between labels. In order to capture the correlations between labels, the sequence-to-sequence (Seq2Seq) model views the MLTC task as a sequence generation problem, which achieves excellent performance on this task. However, the Seq2Seq model is not suitable for the MLTC task in essence. The reason is that it requires humans to predefine the order of the output labels, while some of the output labels in the MLTC task are essentially an unordered set rather than an ordered sequence. This conflicts with the strict requirement of the Seq2Seq model for the label order. In this paper, we propose a novel sequence-to-set framework utilizing deep reinforcement learning, which not only captures the correlations between labels, but also reduces the dependence on the label order. Extensive experimental results show that our proposed method outperforms the competitive baselines by a large margin. 
\end{abstract}

\section{Introduction}
Multi-label text classification (MLTC) is an important yet challenging task in natural language processing (NLP), which aims to assign multiple labels to each sample in the dataset. It can be applied in many real-world scenarios, such as text categorization~\cite{tc}, tag recommendation~\cite{tr}, information retrieval~\cite{ir}, and so on. 

Early work exploring the MLTC task focuses on traditional machine learning algorithms. For instance, binary relevance (BR) \cite{br} decomposes the MLTC task into independent binary classification problems. However, it ignores the correlations between labels.  Much of the following work, including ML-DT \cite{ml_dt}, Rank-SVM \cite{rank_svm}, LP \cite{lp}, ML-KNN \cite{ml_knn}, and CC~\cite{cc}, strives to model the correlations between labels. However, they are computationally intractable when high-order label correlations are considered.

Recent studies turn to deep neural networks, which have achieved great success in the field of NLP. Although they demonstrate a certain degree of improvements, most neural network models~\cite{bp_mll,nn,haram,initial} do not capture the high-order correlations between labels very well. \newcite{YangCOLING2018} propose to apply the Seq2Seq model with attention mechanism to address the MLTC task, which achieves excellent performance. The high-order correlations between labels are well captured through the powerful ability of recurrent neural network (RNN) to model sequence dependencies.

However, the Seq2Seq model is not suitable for the MLTC task in essence. The Seq2Seq model is trained with the maximum likelihood estimation (MLE) method and the cross-entropy loss function, which relies on strict label order. Previous work~\cite{vinyals2015} proves that the order has a great impact on the performance of the Seq2Seq model, which is also verified in our experiments.
Thus, the order of the output labels needs to be predefined carefully. It is reasonable to apply the Seq2Seq model to the MLTC task only when there exists a \emph{perfect label order}. The \emph{perfect label order} means that there is a strict order in the output labels and this true label order is known in practice. However, the \emph{perfect label order} is usually unavailable for the following reasons:

\begin{itemize}
    \item Some labels are naturally unordered. Imposing order to these labels is unreasonable.
    \item Even though there exists a strict order of the output labels, this true label order is usually unknown in practice.
\end{itemize}

It is more appropriate to treat these labels that do not show a \emph{perfect label order} as an unordered set rather than an ordered sequence. An important property of the unordered set is \emph{swapping-invariance}, which means that swapping any two elements in the set will make no difference. This conflicts with the strict requirement of the Seq2Seq model for the label order. 
Therefore, it is inappropriate to directly apply the traditional Seq2Seq model trained with the MLE method to the MLTC task. Otherwise, one of the practical problems that may result is \emph{wrong penalty}. 
\emph{Wrong penalty} means that the model may be wrongly penalized by the MLE method due to inconsistent label order when generating labels that do not show a \emph{perfect label order}. For instance, when the true labels are \{\texttt{A}, \texttt{B}, \texttt{C}\}, the Seq2Seq model still receives a great penalty for generating a label sequence [\texttt{C}, \texttt{A}, \texttt{B}], even though all labels have been predicted correctly.

Although the \emph{perfect label order} is usually unavailable in practice, sometimes the human prior knowledge of the label order can provide valuable information for label prediction to improve the model performance. For instance, when labels are organized in a \emph{directed acyclic graph} (DAG), \newcite{nam2017} take advantage of label hierarchies to place the labels that have same ancestors in the graph next to each other to improve the performance of the Seq2Seq model. However, even if we can grasp the prior knowledge of label order, the Seq2Seq model is still likely to suffer from potential \emph{wrong penalty} because \emph{perfect label order} may not exist. Therefore, an appropriate model for the MLTC task should make use of the human prior knowledge rationally and be free from the strict restriction of the label order. 

Based on this motivation, we propose a novel sequence-to-set model, which can not only integrate human prior knowledge rationally, but also reduce the dependence on the label order. The core component of the proposed model is the bi-decoder structure, which consists of a sequence decoder and a set decoder. The sequence decoder trained by the MLE method is used to fuse human prior knowledge of the label order and the set decoder aims to reduce the dependence of the model on the label order. For the set decoder, we apply the policy gradient method to directly optimize a specific metric that is independent of the label order. Since this specific metric satisfies \emph{swapping-invariance} of the set, the dependence of the model on the label order can be reduced. To the best of our knowledge, this work is the first endeavor to apply reinforcement learning algorithm to the MLTC task.

The contributions of this paper are listed as follows:
\begin{itemize}
	\item We systematically analyze the drawbacks of the current models for the multi-label text classification task.
    \item We propose a novel sequence-to-set model based on deep reinforcement learning, which not only captures the correlations between labels, but also reduces the dependence on the label order.
    \item Extensive experimental results show that our proposed method outperforms the baselines by a large margin. Further analysis demonstrates the effectiveness of the proposed method in addressing the \emph{wrong penalty}.
\end{itemize}

\section{Proposed Approach}\label{methods} 
\subsection{Overview}
Here we define some notations and describe the MLTC task. Given a text sequence $\bm{x}$ containing $m$ words, the task is to assign a subset $\bm{y}$ containing $n$ labels in the label space $\mathcal{L}$ to $\bm{x}$. From the perspective of sequence generation, once the order of the output labels is predefined, the MLTC task can be regarded as the prediction of target label sequence $\bm{y}$ given a source text sequence $\bm{x}$.

\begin{figure}[tb]
	\centering
	\subcaptionbox{Neural sequence-to-set model}{\includegraphics[width=1.0\linewidth]{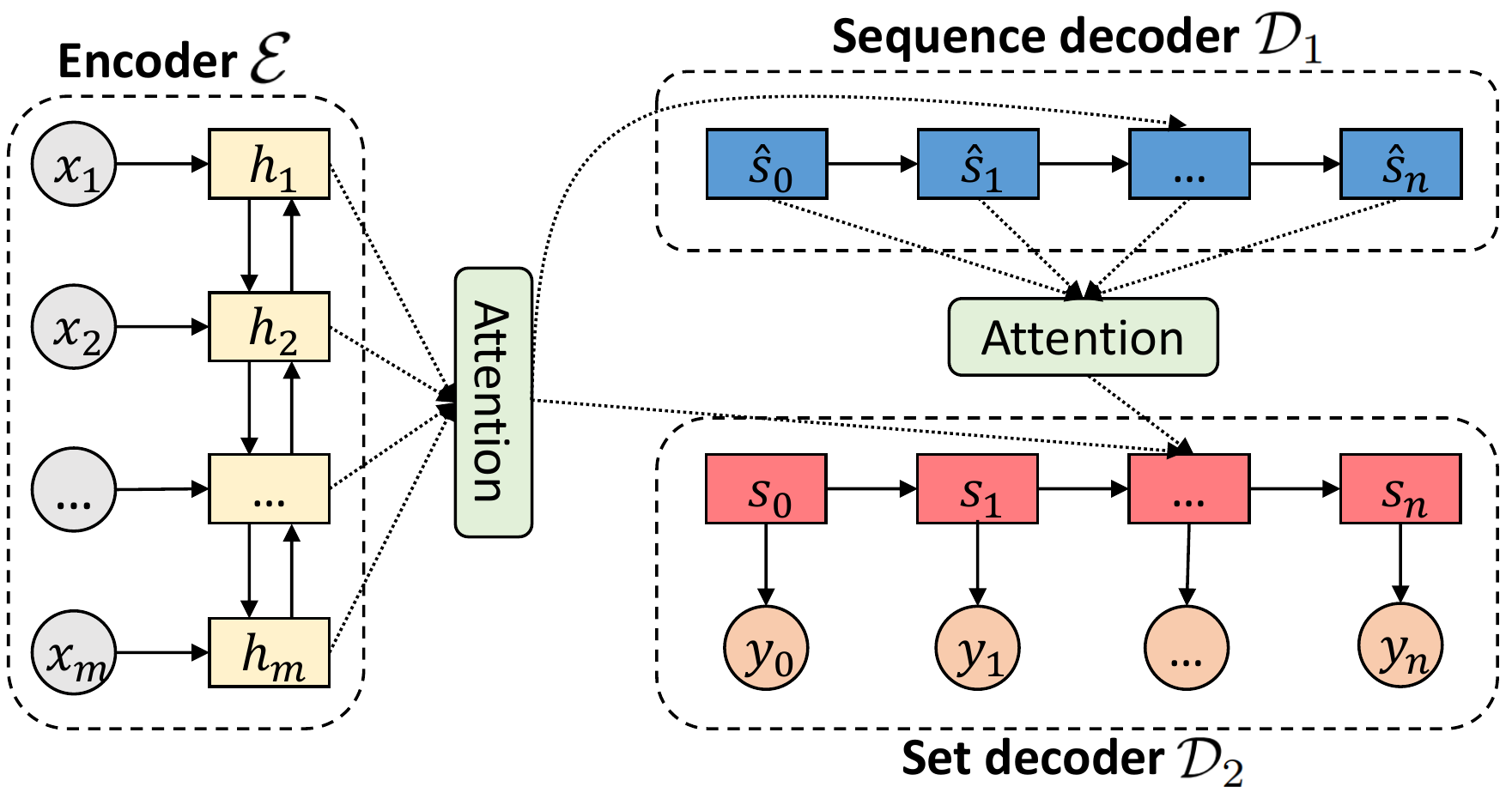}} \\
    \vspace{+0.05in}
	\subcaptionbox{Self-critical training method}{\includegraphics[width=1.0\linewidth]{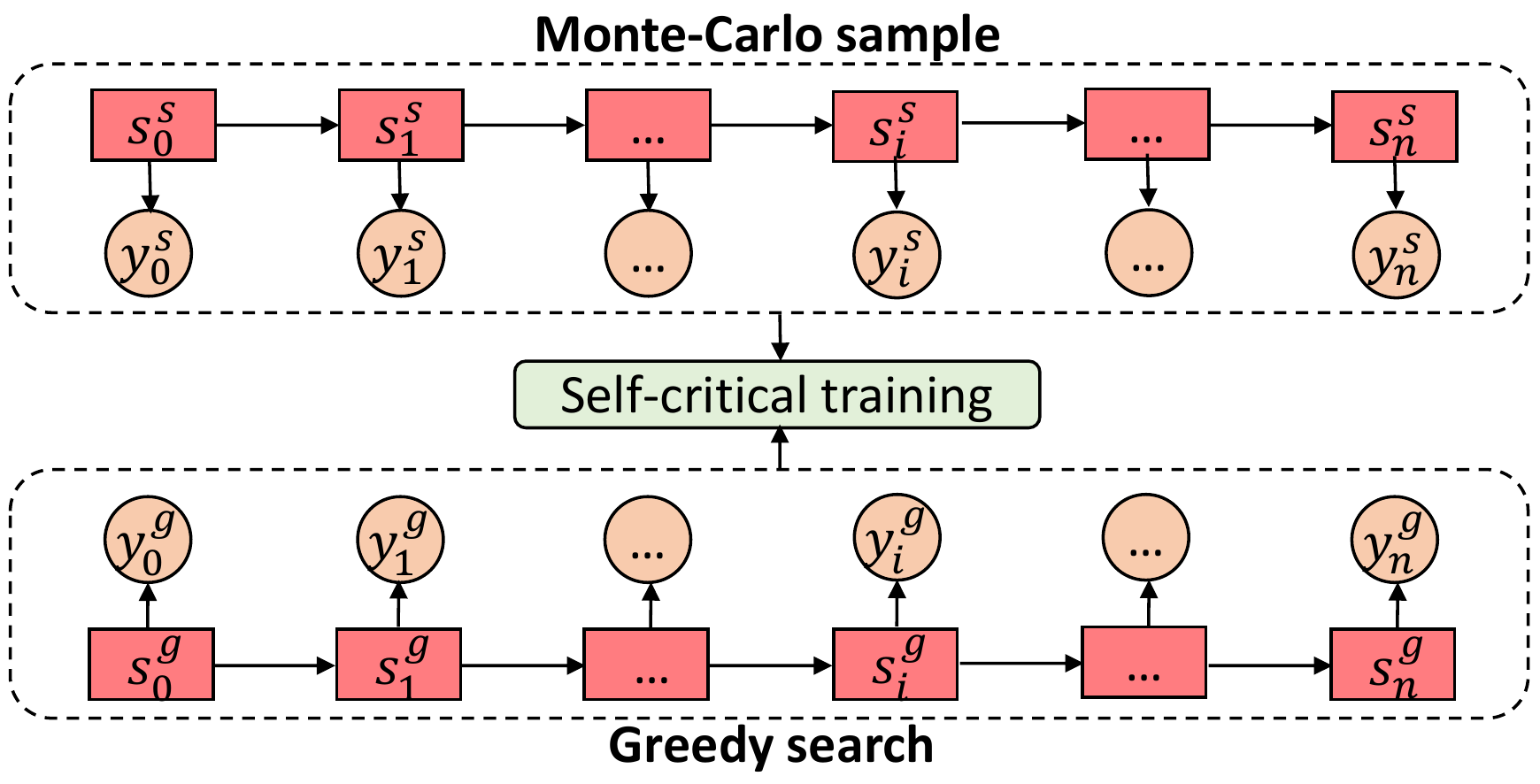}}
	\caption{The overview of our proposed Seq2Set model. Sub-figure (a) shows the architecture of the Seq2Set model and sub-figure (b) shows the self-critical training method. $(y_0^s, \cdots, y_n^s)$ and $(y_0^g, \cdots, y_n^g)$ in (b) represent the sampled label sequence using Monte-Carlo sample and the generated label sequence using greedy search, respectively.}
    \label{model_fig}
\end{figure}

An overview of the proposed model is shown in Figure~\ref{model_fig}. The proposed model consists of an encoder $\mathcal{E}$ and two decoders $\mathcal{D}_1, \mathcal{D}_2$. First, if we have a prior knowledge of the label order, we can utilize it to sort the label sequence to assist the model in achieving better performance. Here we sort labels by frequency in a descending order (from frequent to rare labels)\footnote{In fact, there is a tree structure between the labels of the RCV1-V2 dataset. For instance, the node (label) \texttt{C15} has many child nodes, including \texttt{C151}, \texttt{C152}, etc. The parent nodes appear more frequently. The descending order of label frequency corresponds to the tree structure from the parent node to the child node. This additional information is valuable for label prediction.}. Besides, the $bos$ and $eos$ symbols are added to the head and tail of the label sequence, respectively. Given the input text sequence $\bm{x}$, the encoder $\mathcal{E}$ and the sequence decoder $\mathcal{D}_1$ jointly work like the standard Seq2Seq model to generate corresponding hidden representations $\hat{\bm{s}} = (\hat{s}_0, \hat{s}_1, \cdots, \hat{s}_n)$, which aims to learn a preliminary label order that conforms to possible human prior knowledge. The set decoder $\mathcal{D}_2$ takes both the hidden state vectors of $\mathcal{E}$ and $\mathcal{D}_1$ as input to generate final predicted labels sequentially. The policy gradient method with self-critical training~\cite{rennie2016} is used to reduce the dependence of the model on the label order and alleviate the potential \emph{wrong penalty} problem.

\subsection{Neural Sequence-to-Set Model}
The overview of the proposed neural sequence-to-set model is shown in Figure~\ref{model_fig}(a), which consists of encoder $\mathcal{E}$, sequence decoder $\mathcal{D}_1$ and set decoder $\mathcal{D}_2$.

\textbf{Encoder $\mathcal{E}$:} Here we employ a bidirectional LSTM as the encoder $\mathcal{E}$. The encoder $\mathcal{E}$ reads the input text sequence $\bm{x}$ from both directions and compute the hidden states for each word, 
\begin{align}
\overrightarrow{h}_i & = \overrightarrow{\rm LSTM} (\overrightarrow{h}_{i-1}, x_i) \label{equb_1}\\
\overleftarrow{h}_i & = \overleftarrow{\rm LSTM}(\overleftarrow{h}_{i+1}, x_i) \label{equb_2}
\end{align}

The final hidden representation of the $i$-th word is $h_i = [\overrightarrow{h}_i; \overleftarrow{h}_i]$, which semicolon represents the vector concatenation.

\textbf{Sequence decoder $\mathcal{D}_1$:} Given the hidden state vectors $(h_1, \cdots, h_m)$ of the encoder $\mathcal{E}$, the sequence decoder $\mathcal{D}_1$ performs decoding as the standard Seq2Seq model, which aims to learn a rough label order that conforms to possible human prior knowledge. When $\mathcal{D}_1$ predicts different labels, not all text words make the same contribution. Therefore, the attention mechanism is used to aggregate the hidden representations of informative words. In particular, the hidden state $\hat{s}_{t+1}$ of $\mathcal{D}_1$ at time-step $t+1$ is computed as follows:
\begin{align}
e_{t,i} & = \bm{v}_a^T\tanh(\bm{W}_a\hat{s}_t + \bm{U}_ah_i) \label{equ1} \\
\alpha_{t,i} & = \frac{\exp(e_{t,i})}{\sum_{j=1}^{m}\exp(e_{t,j})} \label{equ2} \\
c_t & = \sum_{i=1}^{m}\alpha_{t,i}h_i \label{equ3} \\
\hat{s}_{t+1} & = {\rm LSTM}(\hat{s}_{t},\left[e(y_{t}); c_{t} \right]) \label{equ4}
\end{align}
where $[e(y_{t}); c_{t}]$ means the concatenation of the vectors $e(y_{t})$ and $c_{t}$. $\bm{v}_a$, $\bm{W}_a$, and $\bm{U}_a$ are weight parameters. $e(y_{t})$ is the embedding of the label which has the highest probability at the last time-step. The probability distribution over the label space is computed as follows:
\begin{align}
o_t & = \bm{W}_of(\bm{W}_d\hat{s}_t + \bm{V}_dc_t) \label{equ5}\\
y_t & \sim softmax(o_t + I_t) \label{equ6}
\end{align}
where $\bm{W}_o$, $\bm{W}_d$, and $\bm{V}_d$ are weight parameters, $I_t \in \mathbb{R}^L$ is the mask vector which is used to prevent the sequence decoder $\mathcal{D}_1$ from generating repeated labels, $f$ is a nonlinear activation function. 
\begin{equation}\label{equ7}
(I_t)_i=
\begin{cases}
- \infty & \text{if the $i$-th label has been predicted.}\\
0 & \text{otherwise.}
\end{cases}
\end{equation}

\textbf{Set decoder $\mathcal{D}_2$:}  The set decoder $\mathcal{D}_2$ is our main model, which aims to capture the correlations between labels and get rid of a strong dependence on the label order. After the sequence decoder $\mathcal{D}_1$ generates the corresponding hidden state sequence, the set decoder $\mathcal{D}_2$ refers to the hidden states of $\mathcal{D}_1$ to generate more accurate labels sequentially. Due to the strong ability of the LSTM structure to model sequence dependencies, the set decoder $\mathcal{D}_2$ is able to capture the correlations between labels. In detail, given $(h_1, \cdots, h_m)$ (the hidden state vectors of the encoder $\mathcal{E}$) and $(\hat{s}_0, \cdots,\hat{s}_n)$ (the hidden state vectors of the sequence decoder $\mathcal{D}_1$), two attention mechanisms similar to Eq.~\ref{equ1} - Eq.~\ref{equ3} aggregate the hidden representations of the encoder and sequence decoder respectively to form the encoder-side context $c_t^e$ and the sequence-decoder-side context $c_t^d$. The hidden state $s_{t+1}$ of the set decoder $\mathcal{D}_2$ at time-step $t+1$ is computed as follows:
\begin{equation}\label{equ8}
s_{t+1} = {\rm LSTM}(s_{t},\left[e(y_{t}); c_{t}^e; c_{t}^d\right])
\end{equation}

The mask vector and softmax layer similar to Eq.~\ref{equ6} - Eq.~\ref{equ7} are used to get the final predicted labels.

\subsection{Training and Testing}
\subsubsection*{Multi-Label Text Classification as a RL Problem}
In order to free the model out of the label order as much as possible, here we model the MLTC task from the perspective of reinforcement learning. Thanks to its ability to directly optimize a specific metric, the policy gradient method has achieved great success in the field of NLP~\cite{ranzato2015,li2016,rennie2016,yu2017}. From the perspective of reinforcement learning, our set decoder $\mathcal{D}_2$ can be viewed as an \emph{agent}, whose \emph{state} at time-step $t$ is the current generated labels $(y_0, \cdots, y_{t-1})$. A stochastic \emph{policy} defined by the parameters $\theta$ of the set decoder $\mathcal{D}_2$ decides the \emph{action}, which is the prediction of the next label. Once a complete label sequence $\bm{y}$ ending with $eos$ is generated, the \emph{agent} $\mathcal{D}_2$ will observe a \emph{reward} $r$. The goal of training is to minimize the negative expected reward: 
\begin{equation}
L(\theta) = - \mathbb{E}_{\bm{y} \sim p_{\theta}}[r(\bm{y})] \label{equ9}
\end{equation}

In our approach, we use the self-critical policy gradient training algorithm~\cite{rennie2016,paulus2017}. The overview of the self-critical training method is shown in Figure~\ref{model_fig}(b). For each training example in the minibatch, the expected gradient of Eq.~\ref{equ9} can be approximated as:
\begin{equation}
\nabla_{\theta}L(\theta) \approx - [r(\bm{y}^s)-r(\bm{y}^g)]\nabla_{\theta}{\rm log}\left(p_{\theta}(\bm{y}^s)\right) \label{equ10}
\end{equation}
where $\bm{y}^s$ is the sampled label sequence from the probability distribution $p_\theta$ and $\bm{y}^g$ is the generated label sequence using the greedy search algorithm. $r(\bm{y}^g)$ in Eq.~\ref{equ10} is the baseline, which is used to reduce the variance of gradient estimate and enhance the consistency of the model training and testing to alleviate \emph{exposure bias} \cite{ranzato2015}.

\subsubsection*{Reward Design}
In order to free the model from the strict restriction of label order and alleviate the potential \emph{wrong penalty}, we give an effective solution based on a simple but intuitive idea: if the reward $r(\bm y)$ can well evaluate the quality of generated label sequence $\bm y$ and satisfy \emph{swapping-invariance} of output labels at the same time, then the dependence of model on the label order can be reduced. Based on this motivation, we design the reward $r$ as the ${\rm F_1}$ score calculated by comparing the output labels with true labels\footnote{When calculating ${\rm F_1}$ score, we convert $\bm y$ and $\bm y^\ast$ into $L$-dimensional sparse vectors.}.
\begin{equation}
r(\bm y) = {\rm F_1}(\bm y, \bm y^\ast) \label{equ11}
\end{equation}
where $\bm y$ and $\bm y^\ast$ are the generated label sequence and the ground-truth label sequence, respectively.

\subsubsection*{Training Objective}
We encourage the sequence decoder $\mathcal{D}_1$ to generate preliminary label sequence whose order conforms to possible human prior knowledge. Therefore, the $\mathcal{D}_1$ is trained by maximizing the conditional likelihood of the ground-truth label sequence $\bm{y}^\ast = (y_0^\ast, \cdots, y_{n}^\ast)$. Specially,
\begin{equation} \label{equ13}
L(\phi) = - \sum_{t=0}^n{\rm log}\left(p(y_t^*|\bm{y}_{<t}^\ast, \bm{x})\right) 
\end{equation}
where $\phi$ represents the parameters of $\mathcal{D}_1$ and $\bm{y}_{<t}^\ast$ denotes the sequence $(y_0^\ast, \cdots, y_{t-1}^\ast)$. The final objective function is as follows:
\begin{equation} \label{equ14}
L_{total} = (1-\lambda) L(\phi) + \lambda L(\theta) 
\end{equation}
In Eq.~\ref{equ14}, $L(\phi)$ aims to fuse the human prior knowledge of the label order into the model and $L(\theta)$ is responsible for reducing the dependence on the label order. $\lambda$ is a hyper-parameter, which is used to control the trade-off between $L(\phi)$ and $L(\theta)$.

\subsection{A Simplified Variant}
Here we provide a simplified variant of the above model. Both the complete model and the simplified model can substantially outperform the previous work and they apply to different scenarios, respectively. There is no clear conclusion as to which of the two proposed models is better. Which model to use depends on the specific task. The complete model is more suitable for the situation where we can grasp a certain amount of prior knowledge of the label order. Otherwise, the simplified model tends to perform better. Detailed analysis is provided in Section \emph{Comparison between Two Proposed Methods}.

In the simplified variant, the sequence decoder $\mathcal{D}_1$ can be removed so that the simplified model only consists of the encoder $\mathcal{E}$ and the set decoder $\mathcal{D}_2$. The simplified model only uses the policy gradient method with self-critical training algorithm for training, and the formula is exactly the same as Eq.~\ref{equ10}.

\begin{table}[t]
\centering
\footnotesize
\setlength{\tabcolsep}{6pt}
\begin{tabular}{l|c c}
  \toprule
   \textbf{Dataset} & \textbf{Words/Sample} & \textbf{Labels/Sample}  \\ \hline
   RCV1-V2 & \num{123.94} & \num{3.24} \\ 
  	AAPD & \num{163.42} & \num{2.41} \\ 
    \bottomrule
    \end{tabular}
    \caption{Statistics of datasets. \textbf{Words/Sample} and \textbf{Labels/Sample} are the average number of words and labels per sample, respectively.}
    \label{tab_datasets}
\end{table}

\section{Experiments}\label{experiments}
In this section, we introduce the datasets, evaluation metrics, the baseline models as well as our experiment settings.

\subsection{Datasets}
\noindent\textbf{Reuters Corpus Volume I (RCV1-V2):} 
The RCV1-V2 dataset is provided by \newcite{rcv1}, which consists of over \num{800000} manually categorized newswire stories. There are 103 topics in total and multiple topics can be assigned to each newswire story.  

\noindent\textbf{Arxiv Academic Paper Dataset (AAPD)
:} This dataset is provided by \newcite{YangCOLING2018}. The dataset consists of the abstract and corresponding subjects of \num{55840} academic papers. The total number of subjects is 54 and each paper can have multiple subjects. The target is to predict subjects of an academic paper according to the content of the abstract. 

For both datasets, we filter out samples with more than 500 words, which removes about 0.5\% of the samples in each dataset. We divide each dataset into training, validation and test sets. The statistic information of the two datasets is shown in Table~\ref{tab_datasets}.

\subsection{Evaluation Metrics}
Following the previous work~\cite{ml_knn,YangCOLING2018}, we adopt hamming loss and micro-${\rm F_1}$ score to evaluate the performance of our models. For reference, the micro-precision as well as micro-recall are also reported.

\begin{itemize}
	\item \noindent\textbf{Hamming-loss} \cite{hamming_loss} is the fraction of labels that are incorrectly predicted.
	\item \noindent\textbf{Micro-${\rm F_1}$} \cite{f1} is the weighted average of ${\rm F_1}$ score of each class.
\end{itemize}

\begin{table}[t]
\centering
\footnotesize
\setlength{\tabcolsep}{6pt}
\begin{tabular}{l|c c c c c}
  \toprule
  \multirow{2}{*}{\textbf{Dataset}} & \textbf{Vocab} & \textbf{Embed} & \textbf{LSTM} & \textbf{Hidden} & \multirow{2}{*}{$\bm\lambda$} \\ 
  & \textbf{Size} & \textbf{Size} & \textbf{Layer} & \textbf{Size} & \\ \hline
   RCV1-V2 & \num{50000} & 256 & (2, 3) & (256, 512) & 0.80 \\ 
  	AAPD & \num{30000} & 256 & (2, 2) & (256, 512) & 0.95  \\ 
    \bottomrule
    \end{tabular}
    \caption{Main experimental hyper-parameters. For the LSTM layer and hidden size, we use $(\cdot, \cdot)$ to represent the hyper-parameter of the encoder and decoder, respectively.}
    \label{tab_parameters}
\end{table}

\subsection{Baselines}
We compare our methods with the following baselines:
\begin{itemize}
	\item \noindent\textbf{Binary Relevance (BR)}~\cite{br} amounts to independently training one binary classifier for each label.
	\item \noindent\textbf{Classifier Chains (CC)}~\cite{cc} transforms the MLTC task into a chain of binary classification problems to model the correlations between labels. 
    \item \noindent\textbf{Label Powerset (LP)}~\cite{lp} creates one binary classifier for every label combination attested in the training set.
    \item \noindent\textbf{CNN}~\cite{cnn} uses multiple convolution kernels to extract text feature, which is then inputted to the linear transformation layer followed by a sigmoid function to output the probability distribution over the label space.
	\item \noindent\textbf{CNN-RNN}~\cite{cnn-rnn} presents an ensemble approach of CNN and RNN to capture both the global and the local textual semantics.
    \item \noindent\textbf{Seq2Seq}~\cite{YangCOLING2018} applies the sequence-to-sequence model~\cite{seq2seq} to perform multi-label text classification.
\end{itemize}

We tune hyper-parameters of all baselines on the validation set based on the micro-$\rm F_1$ score. 

\subsection{Experiment Settings}
We implement our experiments in PyTorch on an NVIDIA 1080Ti GPU. The hyper-parameters of the model on two datasets are shown in the Table~\ref{tab_parameters}. For both datasets, the batch size is set to 64 and out-of-vocabulary (OOV) words are replaced with $unk$. The word embedding is randomly initialized and learned from scratch. We use the Adam~\cite{adam} optimizer to minimize the final objective function. The learning rate is initialized to $0.0003$ and it is halved after every training epoch. Besides, we make use of the dropout method \cite{dropout} to avoid overfitting and clip the gradients~\cite{gradientclip} to the maximum norm of 10. 

During training, we train the model for a fixed number of epochs and monitor its performance on the validation set after 100 updates. Once the training is finished, we select the model with the best micro-$\rm{F_1}$ score on the validation set as our final model and evaluate its performance on the test set.

\section{Results and Discussion}
In this section, we report the results of our experiments on two datasets. For simplicity, we denote our proposed deep reinforced sequence-to-set model as \textbf{Seq2Set}.

\subsection{Results}
The experimental results of our methods and the baselines on the RCV1-V2 dataset are shown in Table~\ref{tab_rcv1}. Results show that both of our proposed methods outperform all baselines by a large margin and the proposed Seq2Set model achieves the best performance in the main evaluation metrics. For instance, our Seq2Set model achieves a reduction of 18.60\% hamming-loss and an improvement of 3.38\% micro-${\rm F_1}$ score over the most commonly used baseline BR. In addition, it also has a large margin over the traditional Seq2Seq model, which shows that the use of reinforcement learning is of great help to improve the accuracy of classification.

\begin{table}[tb]
		\centering
        \footnotesize
    	\setlength{\tabcolsep}{9.0pt}
		\begin{tabular}{l | c c c c }
		\toprule
        \textbf{Models} & \textbf{HL(-)} & \textbf{P(+)} & \textbf{R(+)} & \textbf{F1(+)} \\
		\hline
        BR & 0.0086 & 0.904 & 0.816 & 0.858\\
		CC & 0.0087  & 0.887 & 0.828 & 0.857\\
        LP & 0.0087  & 0.896 & 0.824 & 0.858 \\
        CNN & 0.0089  & \textbf{0.922} & 0.798 & 0.855 \\
		CNN-RNN & 0.0085 & 0.889 & 0.825 & 0.856 \\ 
        Seq2Seq & 0.0081 & 0.889 & 0.848 & 0.868\\ \hline
        \textbf{Seq2Set (simp.)} & 0.0073  & 0.900 & 0.858 & 0.879\\
        \textbf{Seq2Set} & \textbf{0.0070}  & 0.890 & \textbf{0.884} & \textbf{0.887} \\ 
        \bottomrule
		\end{tabular}
		\caption{Performance on the RCV1-V2 test set. HL, P, R, and F1 denote hamming loss, micro-precision, micro-recall and micro-${\rm F_1}$, respectively. The symbol ``+" indicates that the higher the value is, the better the model performs. The symbol ``-" is the opposite. Seq2Set (simp.) denotes the simplified Seq2Set model.}
		\label{tab_rcv1}

	\end{table}

Table~\ref{tab_aapd} presents the experimental results on the AAPD test set. Similar to the results on the RCV1-V2 test set, our proposed methods still outperform all baselines by a large margin. This further shows that our models can not only have significant advantages over other traditional methods, but also reduce the dependence of the model on the label order and alleviate the \emph{wrong penalty}, leading the performance much better than the traditional SeqSeq model.

\subsection{Exploring the Impact of the Label Order}
In order to verify that our models can reduce the dependence on the label order and alleviate the \emph{wrong penalty}, we explore the impact of label order on model performance in two different cases.

\subsubsection*{Uncorrelated Labels}\label{sec_uncorrelated}
In practice, some labels are naturally unordered, for instance, when these labels are uncorrelated. It is not appropriate to use the Seq2Seq model for the MLTC task in this case. To verify that our methods also apply to this case, we rebuilt a new uncorrelated dataset based on the original RCV1-V2 dataset. 

First, we build a collection of uncorrelated labels. The maximum correlation coefficient between these labels is 0.28, indicating that these labels are uncorrelated. Then, we pick out the samples whose labels are all from the uncorrelated label set. We use these samples to rebuild a label-uncorrelated dataset. Table~\ref{uncorrelated} shows the performance of various models on the rebuilt dataset.

According to Table \ref{uncorrelated}, there is no significant difference in the performance of the Seq2Seq model and the BR algorithm on the rebuilt dataset, which shows that the advantage of the Seq2Seq model is not significant in this case. The reason is that the Seq2Seq model makes a strict requirement on the label order, while there is not a strict order between labels in this case, leading that any label permutation will be unreasonable. The Seq2Seq model will suffer from serious \emph{wrong penalty}. However, both of our proposed methods outperform the Seq2Seq model and the BR algorithm by a large margin, which shows that our methods are capable of achieving good performance even in the case where there is not a strict order between labels. Our models are trained using reinforcement learning, and the calculation of reward $r$ satisfies \emph{swapping-invariance}, leading that they reduce the dependence on the label order. This is the main reason for the robustness and universality of our models.

\begin{table}[tb]
		\centering
        \footnotesize
    	\setlength{\tabcolsep}{9.0pt}
		\begin{tabular}{l | c c c c }
		\toprule
        \textbf{Models} & \textbf{HL(-)} & \textbf{P(+)} & \textbf{R(+)} & \textbf{F1(+)} \\
		\hline
		BR & 0.0316 & 0.644 & 0.648 & 0.646 \\
		CC & 0.0306 & 0.657 & 0.651 & 0.654 \\
        LP & 0.0312  & 0.662 & 0.608 & 0.634 \\
        CNN & 0.0256 & \textbf{0.849} & 0.545 & 0.664 \\
		CNN-RNN & 0.0278 & 0.718 & 0.618 & 0.664 \\ 
        Seq2Seq & 0.0255 & 0.743 & 0.646 & 0.691\\ \hline
        \textbf{Seq2Set (simp.)} & \textbf{0.0247}  & 0.739 & \textbf{0.674} & \textbf{0.705} \\
        \textbf{Seq2Set} & 0.0249 & 0.744 & 0.659 & 0.698 \\ 
        \bottomrule
		\end{tabular}
		\caption{Performance on the AAPD test set. Detailed explanations of symbols can be found in Table \ref{tab_rcv1}.}
		\label{tab_aapd}
\end{table}

\begin{figure*}
\begin{minipage}[t]{0.485\linewidth}
\centering
\includegraphics[width=0.85\linewidth]{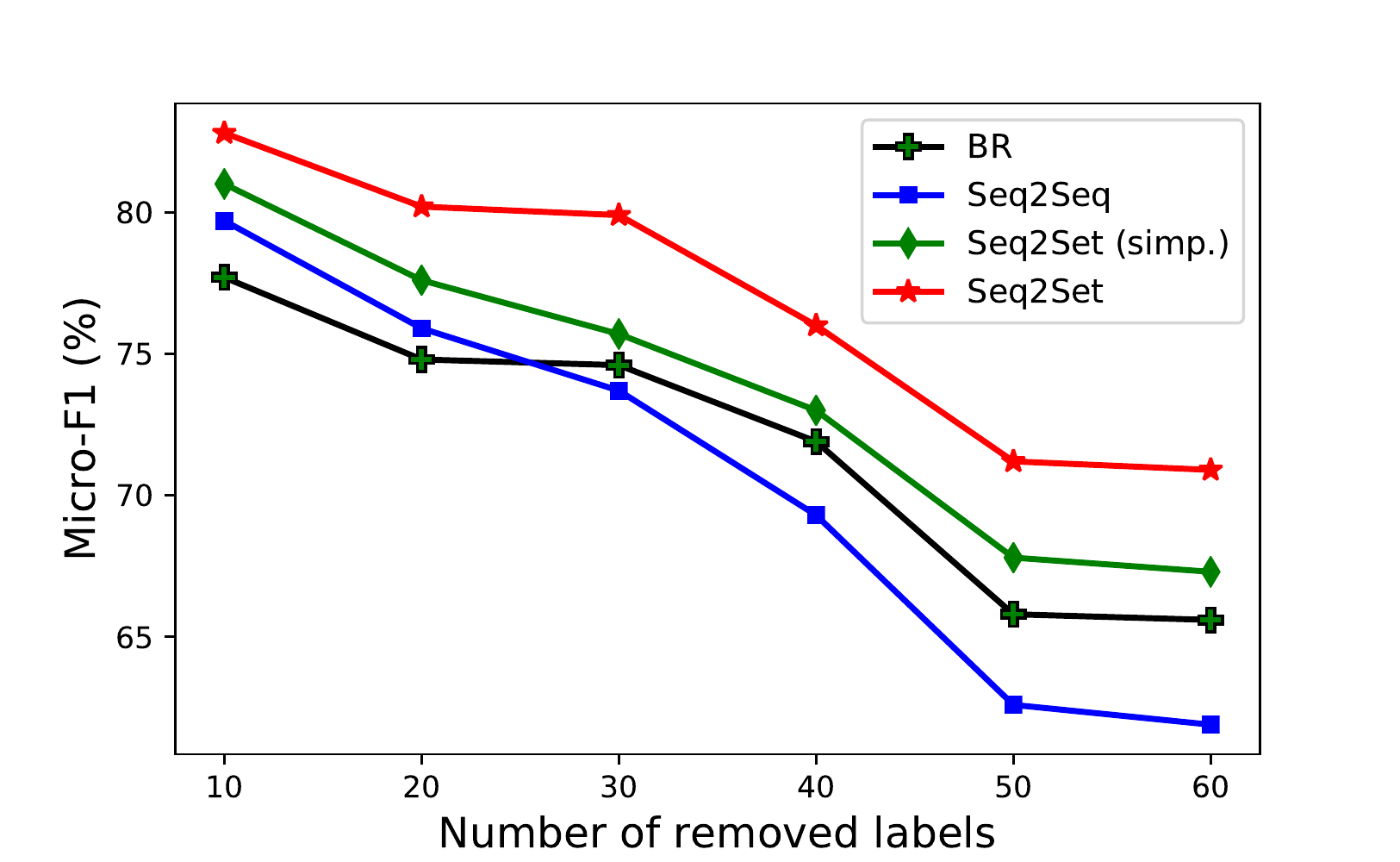}
\caption{Performance of different models when removing different numbers of the most frequent labels. The x-axis refers to the number of the most frequent labels being removed and the y-axis refers to the micro-$\rm {F_1}$ score.}
\label{fig1}
\end{minipage}%
\hspace{0.2in}
\begin{minipage}[t]{0.485\linewidth}
\centering
\includegraphics[width=0.85\linewidth]{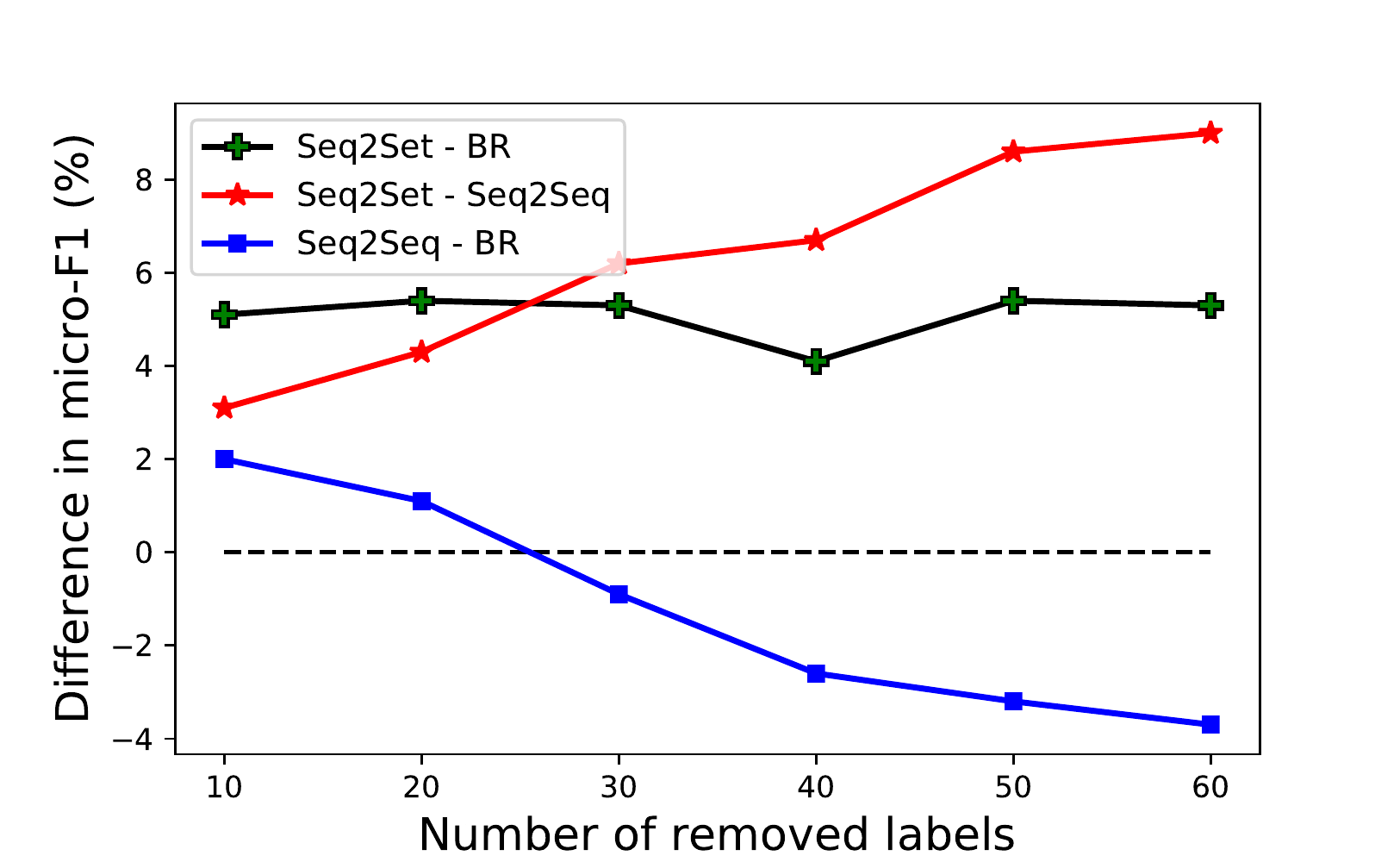}
\caption{The gap of performance of different models when removing different numbers of the most frequent labels. The x-axis refers to the number of the most frequent labels being removed and the y-axis refers to the micro-$\rm {F_1}$ score.}
\label{fig2}
\end{minipage}
\end{figure*}

\begin{table*}
\begin{minipage}[t]{0.485\linewidth}
\centering
\footnotesize
\setlength{\tabcolsep}{9.0pt}
\begin{tabular}{l | c c c c }
\toprule
\textbf{Models} & \textbf{HL(-)} & \textbf{P(+)} & \textbf{R(+)} & \textbf{F1(+)} \\
\hline
BR & 0.0089 & \textbf{0.902} & 0.823 & 0.861 \\
Seq2Seq & 0.0091 & 0.872 & 0.852 & 0.862 \\ \hline 
\textbf{Seq2Set (simp.)} & \textbf{0.0085}  & 0.879 & \textbf{0.862} & \textbf{0.871} \\
\textbf{Seq2Set} & 0.0087 & 0.878 & 0.855 & 0.866 \\ 
\bottomrule
\end{tabular}
\caption{Performance on rebuilt uncorrelated RCV1-V2 test set. Detailed explanations of symbols can be found in Table~\ref{tab_rcv1}.}
\label{uncorrelated}
\end{minipage}%
\hspace{0.2in}
\begin{minipage}[t]{0.485\linewidth}
\centering
\footnotesize
\setlength{\tabcolsep}{9.0pt}
\begin{tabular}{l | c c c c }
\toprule
\textbf{Models} & \textbf{HL(-)} & \textbf{P(+)} & \textbf{R(+)} & \textbf{F1(+)} \\
\hline
BR & 0.0086 & 0.904 & 0.816 & 0.858\\
Seq2Seq & 0.0084 & 0.913 & 0.811 & 0.859 \\ \hline
\textbf{Seq2Set (simp.)} & \textbf{0.0078} & 0.894 & \textbf{0.852} & \textbf{0.873} \\
\textbf{Seq2Set} & 0.0082  & \textbf{0.929} & 0.809 &  0.865 \\ 
\bottomrule
\end{tabular}
\caption{Performance on the label-shuffled RCV1-V2 test set. Detailed explanations of symbols can be found in Table \ref{tab_rcv1}.}
\label{shuffled}
\end{minipage}
\end{table*}

\subsubsection*{Shuffled Labels}
In many real-world scenarios, even if there is a strict order between the labels, this true label order is unknown. We simulate this situation by shuffling the order of the label sequence, which represents that the true label order is unknown. The performance of various models on the label-shuffled RCV1-V2 dataset is shown in Table \ref{shuffled}.

As is shown in Table \ref{shuffled}, when the order of the output labels is shuffled, the performance of the BR algorithm is not affected, but the performance of the Seq2Seq model declines drastically. The reason is that the Seq2Seq model heavily depends on the order of the label sequence, but the labels in the training data present an unordered state. This makes the \emph{wrong penalty} particularly serious, leading to a sharp decline in the performance of the Seq2Seq model. 

However, even in this situation, both of our proposed models are still able to outperform the BR method and the Seq2Seq model by a large margin. For instance, the proposed simplified Seq2Set model achieves a reduction of 7.14\% hamming-loss and an improvement of 1.63\% micro-${\rm F_1}$ score over the Seq2Seq model, which shows that our models have excellent performance regardless of the label order. Our methods reduce the dependence on the label order through reinforcement learning, making them more robust and universal.

\subsection{Comparison between Two Proposed Methods} \label{comparison}

\begin{figure*}[tb]
	\centering
	\includegraphics[width=0.9\linewidth]{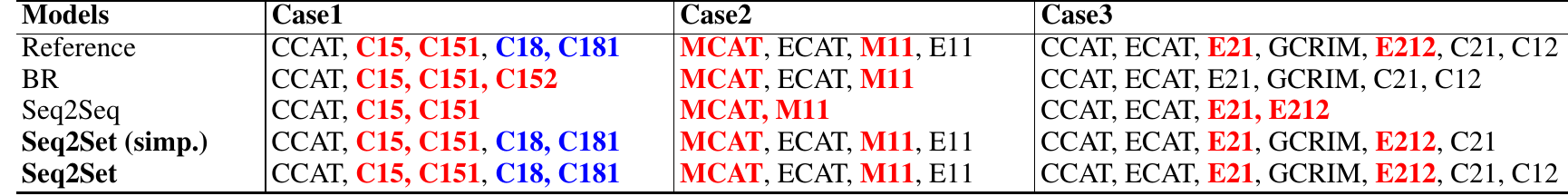}
	\caption{Examples of the generated labels. The labels of the same color in each case are the highly correlated labels.
    }
    \label{case}
\end{figure*}

We can note that there is no clear conclusion as to which of the two proposed models is better. For instance, Table~\ref{tab_rcv1} shows that the proposed Seq2Set model performs better. However, the simplified Seq2Set model achieves the better performance in Table~\ref{uncorrelated} and Table~\ref{shuffled}. The reason is that in the cases of Table~\ref{uncorrelated} and Table~\ref{shuffled}, the labels of training data appear out of order. We do not use valuable prior knowledge to pre-sort the label sequence. Unordered label sequence can cause the sequence decoder $\mathcal{D}_1$ to suffer from the \emph{wrong penalty} to some extent, resulting in the advantage of the Seq2Set model is not significant. Therefore, it is more suitable to use the simplified Seq2Set model in the case where we do not have enough prior knowledge of the label order.

However, if we can grasp a certain amount of prior knowledge of the label order, the Seq2Set model will perform better, which is the corresponding case in Table~\ref{tab_rcv1}. In the case of Table~\ref{tab_rcv1}, we pre-sort the label sequence based on the tree structure of the label set, which is valuable for label prediction. The sequence decoder $\mathcal{D}_1$ encourages the generation of label sequences that conform to the human prior knowledge, which assists the model to achieve better performance. Therefore, the Seq2Set model performs better in this case.

\subsection{Error Analysis}
We find that all methods perform poorly when predicting low-frequency (LF) labels. This is reasonable because the samples assigned LF labels are sparse, making it hard for the model to learn an effective pattern to predict these LF labels. Besides, we also find that the Seq2Seq model exhibits particularly poor performance when predicting LF labels. In fact, as the label set of the RCV1-V2 dataset shows a long-tail distribution, the empirical method of sorting labels from high-frequency to low-frequency is reasonable. However, the distribution of LF labels is relatively uniform, leading that there is no obvious order between these LF labels, which conflicts with the strict requirements of the Seq2Seq model for the label order. Therefore, the Seq2Seq model performs poorly when predicting LF labels. However, in many real-word scenarios of multi-label text classification, the label set does not show a long-tail distribution, which poses a serious challenge to the Seq2Seq model.

In contrast, since our proposed Seq2Set model reduces the dependence on the label order, we believe that it is more universal and robust to different label distributions. In order to verify our conjecture, we remove the top $10, 20, \cdots, 60$ most frequent labels in turn and evaluate the performance of various models on the classification of the remaining labels. The more the most frequent labels are removed, the closer the label distribution is to the uniform distribution. Figure~\ref{fig1} shows changes in the performance of various methods on the RCV1-V2 test set when removing different numbers of the most frequent labels and Figure~\ref{fig2} shows changes in the gap of the performance of different methods.

As is shown in Figure~\ref{fig1}, although the performance of all models deteriorates when the number of the most frequent labels being removed increases, the performance of the Seq2Seq model significantly decreases compared with other methods. However, according to Figure~\ref{fig2}, as more labels are removed, the advantage of our Seq2Set model over the Seq2Seq model gradually becomes larger. At the same time, the advantage of our Seq2Set model does not decrease compared with the BR algorithm. This shows that our proposed methods are more robust and universal, which apply to different label distributions. The reason is that our methods not only capture the correlations between labels, but also reduce the dependence on label order through reinforcement learning.

\subsection{Case Study}
Several examples of the generated label sequences using different models on the RCV1-V2 dataset are shown in Figure~\ref{case}. Due to the tree structure between the labels of the RCV1-V2 dataset, some labels, such as \texttt{C15}, \texttt{C151}, etc., are highly correlated, leading that they are likely to appear together in the predicted labels. The BR algorithm fails to capture this correlation. For example, the true label \texttt{E212} is missed in case3. However, although the Seq2Seq model is able to capture this correlation, it makes a strong dependence on the label order, leading to premature termination of the label sequence or occurrence of some true uncorrelated labels being left out. For instance, the Seq2Seq model misses the true labels \texttt{C18} and \texttt{C181} due to premature termination of the label sequence in case1 and it also misses the true label \texttt{GCRIM} due to the weak correlation between \texttt{GCRIM} and other true labels in case3.

However, our methods can overcome the above mentioned drawbacks of the BR algorithm and Seq2Seq model, which makes the prediction of labels more accurate. For instance, our proposed methods almost accurately predict all true labels in three examples. Our methods predict the next label based on its previously predicted labels by LSTM structure. Therefore, benefiting from the ability of the LSTM structure to model sequence dependencies, they can capture the correlations between labels well. Besides, our methods are trained using reinforcement learning, where the calculation of reward $r$ satisfies \emph{swapping-invariance}. Therefore, they can reduce requirements on the label order, and effectively alleviate the \emph{wrong penalty}, leading to more robustness and universality.

\section{Related Work}
\subsection{Multi-Label Text Classification}
Multi-label text classification (MLTC) aims to assign multiple labels to each sample in the dataset. 
Early work on exploring the MLTC task focuses on machine learning algorithms, mainly including problem transformation methods and algorithm adaptation methods. Problem transformation methods, such as BR~\cite{br}, LP~\cite{lp} and CC~\cite{cc}, map the MLTC task into multiple single-label learning tasks. Algorithm adaptation methods extend specific learning algorithms to handle multi-label data directly. The corresponding representative work is ML-DT~\cite{ml_dt}, Rank-SVM~\cite{rank_svm}, ML-KNN~\cite{ml_knn}, and so on. In addition, some other methods, such as ensemble methods~\cite{rakel}, joint training methods~\cite{li2015multi}, etc., are also used for the MLTC task. However, they can only be used to capture the first or second order label correlations, or are computationally intractable when high-order label correlations are considered.

Recent years, some neural network models have also been successfully used for the MLTC task. For instance, the BP-MLL proposed by~\newcite{bp_mll} applies a fully-connected network and the pairwise ranking loss to perform classification. \newcite{nn} 
further replace the pairwise ranking loss with cross-entropy loss function. \newcite{embed} present an initialization method to model label correlations by leveraging neurons. \newcite{cnn-rnn} present an ensemble approach of CNN and RNN so as to capture both global and local semantic information. The two milestones are~\newcite{nam2017}  and~\newcite{YangCOLING2018}, both of which utilize the Seq2Seq model to capture the label correlations. The former explores the impact of different label permutations on the performance of the model and the latter adds the global embedding to alleviate \emph{exposure bias}. In parallel to our work, \newcite{lin2018} propose a semantic-unit-based dilated convolution model for the MLTC task and \newcite{semene_pred} apply a label distributed Seq2Seq model to learn semantic knowledge, which is a specific application of the MLTC task. 

\subsection{Reinforcement Learning}
Reinforcement learning (RL) explores how an agent should take actions to interact with the environment so as to maximize cumulative reward. It is a long-standing problem, here we focus on its application in natural language processing. \newcite{ranzato2015} propose to train the Seq2Seq model using policy gradient method for machine translation and \newcite{bahdanau2016actor} present an approach to training neural networks to generate sequence using the actor-critic method. \newcite{rennie2016} design the self-critical training method on image captioning task to further alleviate the \emph{exposure bias}. \newcite{li2016} apply deep reinforcement learning to model future reward in chatbot dialogue. SeqGAN~\cite{yu2017} adapts the generative adversarial networks~\cite{goodfellow2014generative} to the task of text generation by treating the procedure of generation as a stochastic policy. Recently, \newcite{wang18} integrate local and global decision-making using reinforcement learning models to solve Chinese zero pronoun resolution. However, as far as we know, there is no work to directly apply the reinforcement learning algorithm to the MLTC task.

\section{Conclusion}\label{conclusion}
In this paper, we systematically analyze the drawbacks of the current models for the multi-label text classification task. A novel sequence-to-set model based on deep reinforcement learning is proposed to fuse the human prior knowledge rationally and reduce the dependence on the label order. Extensive experimental results show that the proposed method outperforms the competitive baselines by a large margin. Further analysis of experimental results demonstrates that our approach not only captures the correlations between labels, but also is free from the strict restriction of the label order, leading to better robustness and universality.

\bibliography{aaai2019}
\bibliographystyle{aaai}

\end{document}